\documentclass[journal]{IEEEtran}
\usepackage{graphicx}
\usepackage{epstopdf}
\usepackage{subfigure}
\usepackage[noend]{algpseudocode}
\usepackage{algorithmicx,algorithm}
\usepackage{amsmath}
\usepackage{amssymb}
\usepackage{amsthm}
\usepackage{xspace}
\usepackage{color}
\usepackage[table]{xcolor}
\usepackage{tikz}
\usetikzlibrary{patterns}
\usepackage{amssymb}
\usepackage{xcolor}
\usepackage{listings}
\usepackage[flushleft]{threeparttable}
\usepackage{pgfplots}
\pgfplotsset{compat=1.3}%
\usepackage{verbatim}
\usepackage{epstopdf}
\usepackage{cite}

\newtheorem{remark}{\textbf{Remark}}

\newtheorem{definition}{Definition}
\newtheorem{theorem}{Theorem}

\newcommand{\eg}{\emph{e.g., }}
\newcommand{\etal}{\emph{et al.}}
\newcommand{\systemname}{CoUAV\xspace}

\newcommand{\nop}[1]{}

\begin{document}

\title{\sc{\systemname}: A Cooperative UAV Fleet Control and Monitoring Platform}

\author{Weiwei~Wu,~\IEEEmembership{Member,~IEEE},
Ziyao~Huang,
Feng~Shan,~\IEEEmembership{Member,~IEEE},
Yuxin~Bian,
Kejie~Lu,~\IEEEmembership{Senior Member,~IEEE},
Zhengjiang~Li,~\IEEEmembership{Member,~IEEE},
Jianping~Wang,~\IEEEmembership{Member,~IEEE},
\IEEEcompsocitemizethanks{
\IEEEcompsocthanksitem W. Wu, Z. Huang, F. Shan, Y. Bian and H. Li are with School of Computer Science and Engineering, Southeast University, Nanjing, Jiangsu, P. R. China. Emails: weiweiwu@seu.edu.cn, seuhuangziyao@outlook.com, shanfeng@seu.edu.cn, byxshr@163.com.
\IEEEcompsocthanksitem K. Lu is with the Department of Computer Science and Engineering, University of Puerto Rico at Mayag\"uez, Puerto Rico. Email: kejie.lu@upr.edu.
\IEEEcompsocthanksitem Z. Li and J. Wang are with Department of Computer Science, City University of Hong Kong, Hong Kong, P.~R.~China. Emails: \{zhenjiang.li, jianwang\}@cityu.edu.hk.
\IEEEcompsocthanksitem The work is supported in part by Science Technology and Innovation Committee of Shenzhen Municipality Under project JCYJ20170818095109386, National Natural Science Foundation of China under Grant No. 61672154, and Aeronautical Science Foundation of China Under Grant No. 2017ZC69011.
}
}

\maketitle

\begin{abstract}
In the past decade, unmanned aerial vehicles (UAVs) have been widely used in various civilian applications, most of which only require a single UAV. In the near future, it is expected that more and more applications will be enabled by the cooperation of multiple UAVs. To facilitate such applications, it is desirable to utilize a general control platform for cooperative UAVs. However, existing open-source control platforms cannot fulfill such a demand because (1) they only support the leader-follower mode, which limits the design options for fleet control, (2) existing platforms can support only certain UAVs and thus lack of compatibility, and (3) these platforms cannot accurately simulate a flight mission, which may cause a big gap between simulation and real flight. To address these issues, we propose a general control and monitoring platform for cooperative UAV fleet, namely, \emph{\systemname}, which provides a set of core cooperation services of UAVs, including synchronization, connectivity management, path planning, energy simulation, etc. To verify the applicability of \systemname, we design and develop a prototype and we use the new system to perform an emergency search application that aims to complete a task with the minimum flying time. To achieve this goal, we design and implement a path planning service that takes both the UAV network connectivity and coverage into consideration so as to maximize the efficiency of a fleet. Experimental results by both simulation and field test demonstrate that the proposed system is viable. 
\end{abstract}


\IEEEpeerreviewmaketitle

\section{Introduction}
In recent years, unmanned aerial vehicles (UAVs), especially multirotor-based drones, have attracted significant attention from federal agencies, industry, and academia. 
Although many existing UAV applications are based on a single UAV, better applications can be facilitated by using multiple cooperative UAVs \cite{Gupta2016, ANJBE-NETW18}. For example, in a video surveillance application, multiple UAVs can quickly scan a given area, and can also improve the performance of scanning using advanced video processing technologies \cite{Meng2015}. 
Nevertheless, to successfully deploy a multi-UAV application and enable the cooperation among UAVs, many challenging issues must be solved, such as flight control, mobility, routing, reliability, safety, etc. \cite{Gupta2016}.

\begin{figure}
    \centering
    \includegraphics[width=0.8\columnwidth]{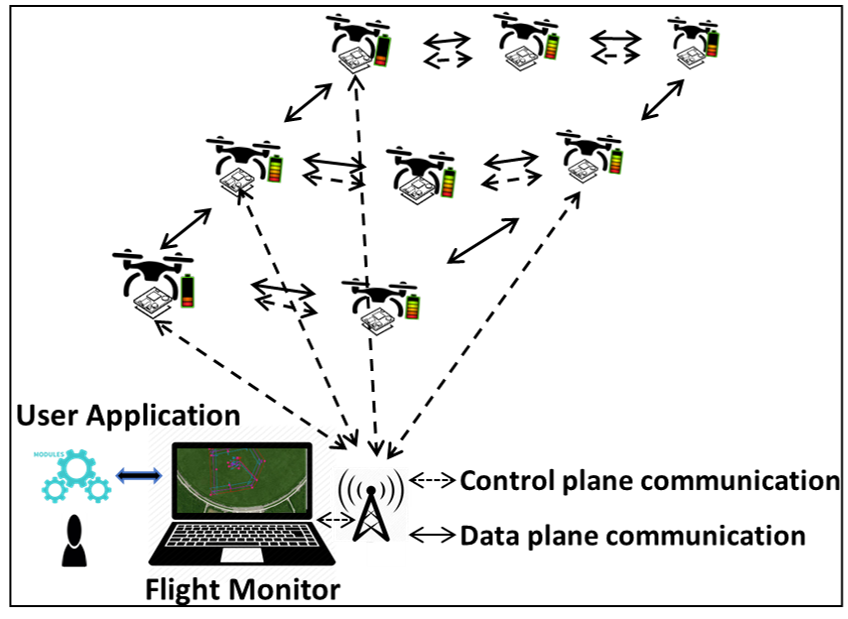}
    \caption{Architecture of the UAV fleet platform.}
    \label{fig_system}
\end{figure}

Clearly, to facilitate multi-UAV applications, it is desirable to utilize a general platform to control and monitor UAVs, as illustrated in Fig.~\ref{fig_system}. In the literature, there exist some open-source control platforms for UAVs, referred as \emph{Ground Control Stations (GCSs)}, including Mission Planner~\cite{mispla},  QGroundControl~\cite{qgdctl} and DJI FLIGHTHUB~\cite{djiflthb}.
Although all of these GCSs support the basic flight control functionality, such as flight planning by editing waypoints, communication with UAVs, user-friendly GUIs, flight trajectory displaying on a map and real-time vehicle status monitoring, 
the following limitations limit their applicability as general control and monitoring platforms for cooperative UAVs.
\begin{table*}[t]
\centering
\caption{Comparison of functionality among CoUAV and popular GCSs}
\begin{threeparttable}
\begin{tabular}{|l|cccccc|}
\hline
                         & Mission Planner & MAVProxy     & DJI FLIGHTHUB & QGroundControl & APM Planner 2 & CoUAV        \\ \hline
GUI                      & $\checkmark$    & $\times$     & $\checkmark$  & $\checkmark$   & $\checkmark$  & $\checkmark$ \\ \hline
Single-UAV Simulation    & $\checkmark$    & $\checkmark$ & $\checkmark$  & $\checkmark$   & $\checkmark$  & $\checkmark$ \\ \hline
UAV APIs\tnote{1}        & $\checkmark$    & $\checkmark$ & $\times$      & $\times$       & $\times$      & $\checkmark$ \\ \hline
GUI APIs\tnote{2}        & $\times$        & $\times$     & $\times$      & $\times$       & $\times$      & $\checkmark$ \\ \hline
Multi-UAV Simulation     & $\times$        & $\times$     & $\times$      & $\times$       & $\times$      & $\checkmark$ \\ \hline
Swarm APIs\tnote{3}      & $\times$        & $\times$     & $\times$      & $\times$       & $\times$      & $\checkmark$ \\ \hline
Hadware-independent      & $\times$        & $\times$     & $\times$      & $\times$       & $\times$      & $\checkmark$ \\ \hline
Energy-consumption model & $\times$        & $\times$     & $\times$      & $\times$       & $\times$      & $\checkmark$ \\ \hline
\end{tabular}
\begin{tablenotes}
  \item[1] Interfaces that enable the developer to communicate with both UAV and the platform.
  \item[2] Functions that help developer fast create and easily control unified-style GUI in the original desktop application to interact with users and handle the input.  
  \item[3] Functions that facilitate the cooperation of UAVs like \lstinline{sendSyncMsg(ID, info, callback)}, \lstinline{ping(ID, callback)} etc.
\end{tablenotes}
\end{threeparttable}
\label{gcs_features}
\end{table*}
\begin{itemize}

\item
Only leader-follower mode is enabled in the existing GCSs. Though leader-follower mode makes path planning much easier, it cannot fully utilize all UAVs in a fleet to complete a complex task at the earliest time or with shortest flying distance.  

\item
Each GCS only supports a specific set of UAVs or UAV flight controllers.
For example, Mission Planner is designed primarily for ArduPilot hardwares and firmwares; QGroundControl only supports UAVs that communicate using the MAVLink protocol; the DJI FLIGHTHUB interacts with DJI's own products only.

\item
There is a lack of energy simulation module in existing GCSs. Without energy simulation module, existing GCSs cannot predict energy consumption through simulation. Thus, the feasibility of a flight cannot be tested before UAVs take off. As a result, some flights may have to be aborted before their tasks are completed due to early energy depletion. 
\end{itemize}

\iffalse
\begin{figure}
    \centering
    \includegraphics[width=1.05\columnwidth]{Pic/couav-general-structure.png}
    \caption{General structure of the program while implementing a multi-UAV algorithm}
    \label{fig:general_structure}
\end{figure}\else

To this end,  we propose \systemname, which is a control and monitor platform that enables easy-to-implement UAV cooperation,  to address the aforementioned limitations. 
Specifically, to address the first limitation and allow UAVs in a fleet to maximize the fleet efficiency, we propose a more generic path planning framework where UAVs do not need to follow leader-follower mode. Instead, the proposed   framework enables cooperative path planning by introducing swarm  functions (e.g., synchronizations, connectivity maintenance). To  demonstrate the functionalities, we provide an embedded path  planning  service  for  the  multi-UAV  cooperation  by considering both the UAV network connectivity and coverage.

To address the second limitation, we provide the hardware-independence to each UAV by introducing a companion linux-kernel device, which serves as a middelware to interact with UAV autopilots. Since almost every commodity provider and open-source community offers linux-based SDK for UAV flight control, such UAV companion devices hide the hardware and software difference of UAVs from different manufacturers. Hence, our \systemname platform is generic enough to work with various UAVs, regardless their hardwares, firmwares, and communication protocols.

To address the third limitation, we add an energy simulation module to the \systemname platform. To make the simulation reliable and close to the real-word flight, we make efforts to energy prediction, which can avoid the task abortions in the field. In a fleet with heterogeneous drones, different UAVs may consume different amount of energy even when they fly at the same speed/cover the same distance. Our platform provides an accurate energy model tailored to different types of UAVs, ensuring the feasibility of the  real flight under planned paths. 

The key differences of functionalities among CoUAV and popular GCSs are summarized in Table~\ref{gcs_features}. Besides the aforementioned functionalities, CoUAV also offers other features such as GUI APIs, UAV APIs, and simulation for multiple UAVs, based on which we implement some basic modules, such as agent manager, emergency monitor, and message center, for ease of application development. A developer can use our platform to achieve rapid development without having to implement the underlying modules.


Our contributions can be summarized as follows.

\begin{itemize}
\item 
\systemname has the advantage to provide effective cooperation services and manage sophisticated networking protocols. The UAV agent developed in \systemname includes an independent middleware that can run a general operating system, on which many open source projects can be executed. This implies that \systemname can support  not only existing mainstream protocols, but also any specialized airborne communication protocols proposed and developed in the future.

\item 
In addition to the primary cooperation services, \systemname can further support sophisticated path planning for a fleet, e.g., connectivity maintenance and synchronization during the flight. 
By taking points to visit and UAV number as input, \systemname can generate the initial path plan so that the task can be completed at the earliest time while maintaining the connectivity among UAVs. The planned path information is then converted to a series of control commands and disseminated to individual UAVs. When the path is impaired due to environmental factors, like wind disturbance, the planned path can be revised and updated.

\item
\systemname provides interfaces to incorporate trained energy models as well as modules to train energy models for different types of UAVs.  By collecting energy data through historic flying tasks, we have learned an energy model for Pixhawk-Hexa UAVs. Comparing the simulation results with the field test results, the training energy model can achieve 94.26\% accuracy.

\item 
\systemname can accurately simulate a flight mission. Moreover, \systemname supports an easy switch between simulations and testbed experiments executing the same task. These advantages come from the system design, where the UAV agent serves as a middleware between the original UAV and the ground station of the \systemname platform to hide the hardware difference. As a result, we can replace any UAV models without affecting other parts of the platform, and also use the UAV simulator to conduct  simulations prior to the deployment. A demo and the source code of the platform are  available for public access in \textit{ https://github.com/whxru/CoUAV}. 

\end{itemize}

The rest of the paper is organized as follows. 
In Section~\ref{sec:design}, we elaborate on the design and implementation of a prototype. To verify the applicability of the proposed system, we provide an efficient path planning service for an emergency search application in Section~\ref{sec:application}, and then conduct extensive simulations and field tests in Section~\ref{sec:results}. Finally, we discuss related work in Section~\ref{sec:related}, before concluding the paper in Section~\ref{sec:conclusion}.

\iffalse
\textcolor{blue}{As shown in Fig.~\ref{fig:general_structure}, we present a general structure of the program while implementing a multi-UAV algorithm and it is divided into three columns from left to right. The first column contains the basic components needed in such a program which also are all provided by CoUAV, the second column contains the APIs published by CoUAV to build the algorithm module in the third column which is the only component developers are going to create by themselves. From bottom to top there are 9 types of basic components:}
\begin{itemize}
\item
\textcolor{blue}{UAV Agent and UAV Controller for each UAV to perform a set of actions, get data from UAV or its onboard sensor(s) and communicate with the platform. Based on them the UAV APIs are published to enable developer to communicate with both UAV and CoUAV platform.}
\item
\textcolor{blue}{Message Center and Communicator for the platform to manage messages and communicate with each UAV Agent. Based on them the Message Register is provided to extend the communication protocol and message types, which make the members in developer's swarm are able to speak in their own language and talk about their own topic.}
\item
\textcolor{blue}{Agent Manager, Energy-consumption Monitor, Emergency Monitor, Core Variable Manager receive related and handled messages from the message center and respectively manage connections of every UAV agent, monitor their energy-consumptions, handle possible emergencies and maintain core variables related to the state of program and swarm. Based on them Swarm APIs are published to facilitate the cooperation of UAVs in a high level.}
\item
\textcolor{blue}{GUI to display the state of swarm, information of task,  warnings and interact with users efficiently. CoUAV provides GUI APIs that help developers fastly create and easily control unified-style GUI in the original desktop application to interact with users and handle the input. }
\end{itemize}

\textcolor{blue}{There exist some well-known control platforms for UAVs, referred as \emph{Ground Control Stations (GCSs)}, including Mission Planner~\cite{mispla}, MAVProxy~\cite{mavproxy}, DJI FLIGHTHUB~\cite{djiflthb}, QGroundControl~\cite{qgdctl}, APM Planner 2~\cite{apmpln}. Although all of those GCSs support the basic flight control functionality for single UAV like waypoints, most of them are lack of support for swarms. A detailed comparison of functionality among CoUAV and popular GCSs is given in Table~\ref{gcs_features} from where we could see the advantages of CoUAV. }
\begin{itemize}
    \item \textcolor{blue}{GUI APIs provided: Although All of the platforms provide GUI except MAVProxy which is a based on command-line and console, only CoUAV provide related APIs for helping developers create and control GUI in the original desktop application efficiently to interact with users.}
    \item \textcolor{blue}{Ability of predicting enerygy-consumption: Only CoUAV has an energy-consumption model which aims at making the simulation reliable and close to the real-word flight in case of the task abortions in the field. In a fleet with heterogeneous drones, different UAVs may consume different amount of energy even when they fly at the same speed/cover the same distance. Our platform provides an accurate energy model tailored to different types of UAVs, ensuring the feasibility of the  real flight under planned paths.}
    \item \textcolor{blue}{Ability of swarm: All of the platforms are able to do single-UAV simulation, but only CoUAV could do multi-UAV simulation and provide Swarm APIs to enable the cooperation of UAVs. Although a few GCSs can keep multiple UAV connections at the same time, things they could do are very limited. For example, in Mission Planner those UAVs can only fly in leader-follower mode and in QGroundControl the user could only apply one same command to every UAV.}
    \item \textcolor{blue}{Hardware-independent: We introduce a companion linux-kernel device to each UAV, serving as a middelware to interact with UAV autopilots. Since almost every commodity provider and open-source community offers linux-based SDK for UAV flight control, such UAV companion devices hide the hardware and software difference of UAVs from different manufacturers. Hence, our \systemname platform is generic enough to work with various UAVs, regardless their hardwares, firmwares, and communication protocols.}
\end{itemize}

\textcolor{blue}{Besides functionalities listed in Table ~\ref{gcs_features}, to hence the ablity of swarm, we also propose a more generic path planning framework where UAVs do not need to follow leader-follower mode. Instead, the proposed framework enables cooperative path planning and aims to minimize the total flying distance of the whole fleet in order to complete a task while ensuring that each UAV can communicate with at least one UAV in the fleet and at least one UAV can communicate with the ground control station at any moment of a flight.}
\else

\section{\systemname System Design and Implementation}
\label{sec:design}

The \systemname platform consists of two types of components: the \textit{UAV agent} installed in each UAV and the \textit{flight monitor} operating on a ground station, as illustrated in Fig.~\ref{fig_platform}. In this section, we present the implementation details of the UAV agent and the flight monitor on the \systemname platform. 

\begin{figure}
    \centering
    \includegraphics[width=.5\textwidth]{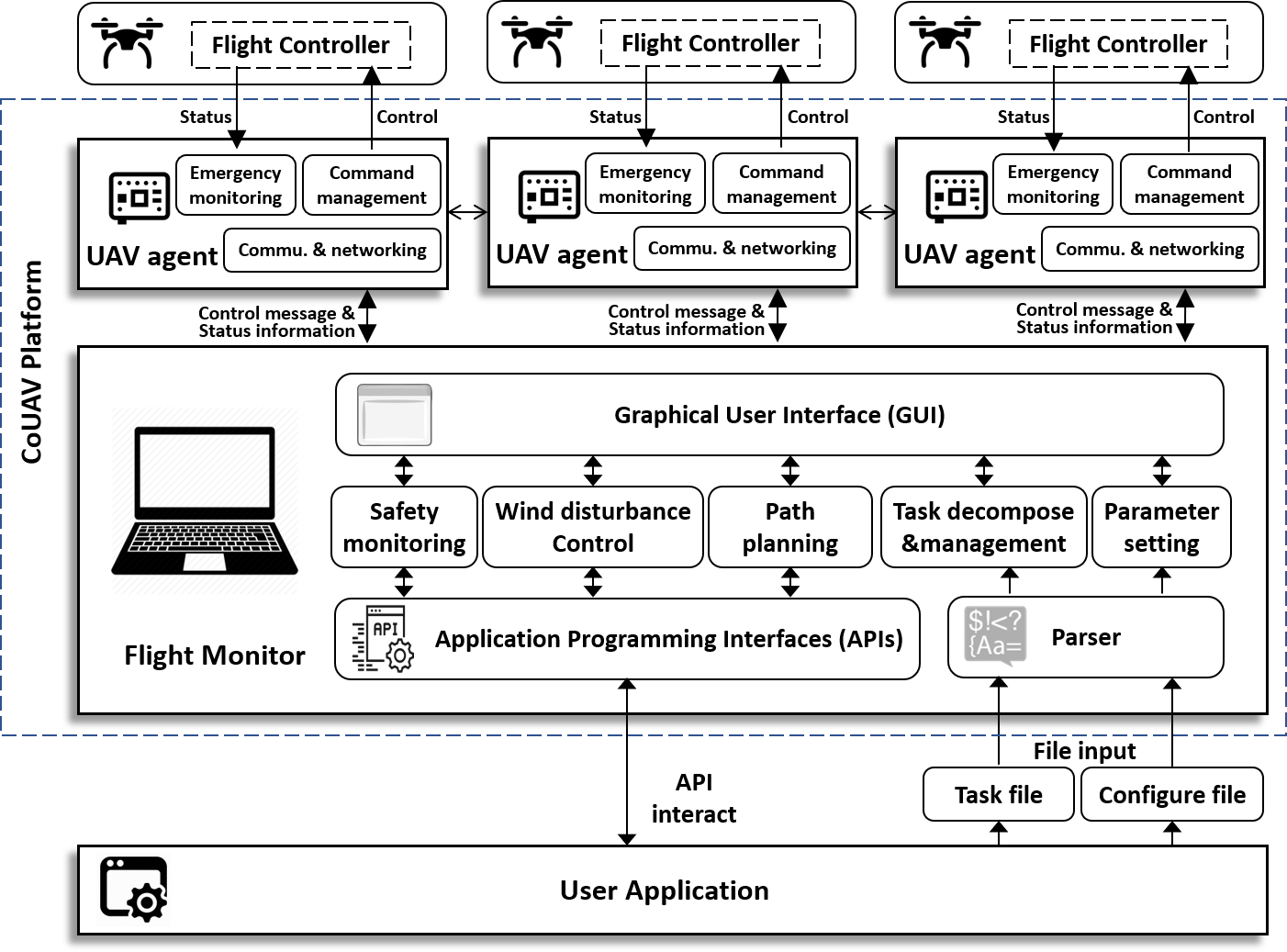}
    \caption{Illustration of our proposed \systemname platform.}
    \label{fig_platform}
\end{figure}

\subsection{The UAV agent}

A typical UAV or drone system consists of motors, flight control system, gyroscope, compass, GPS, remote control and battery.
The main task of the flight control system is to stabilize the vehicle and control its movement through the control of motors, based on the information from gyroscope, compass and GPS.
The flight control system also provides the drone information and control interfaces to external devices by a pre-defined protocol. As shown in Fig. \ref{hardware}, the flight controllers on our current platform are the APM2.8 board and Pixhawk HEXA borad.
We further install a Raspberry Pi 3 motherboard (RPi) as the mounted Linux-kernel device to run the UAV agent program. 
The UAV agent is responsible to handle three important types of information or messages, including vehicle status, device control and exceptions. UAV APIs that communicate between the flight control board and the monitor are provided.
The detailed illustration of the UAV agent and its interaction with the UAV flight controller are illustrated in Fig.~\ref{fig:uavagent}.

\begin{figure}
    \centering
    \includegraphics[width=0.93\columnwidth]{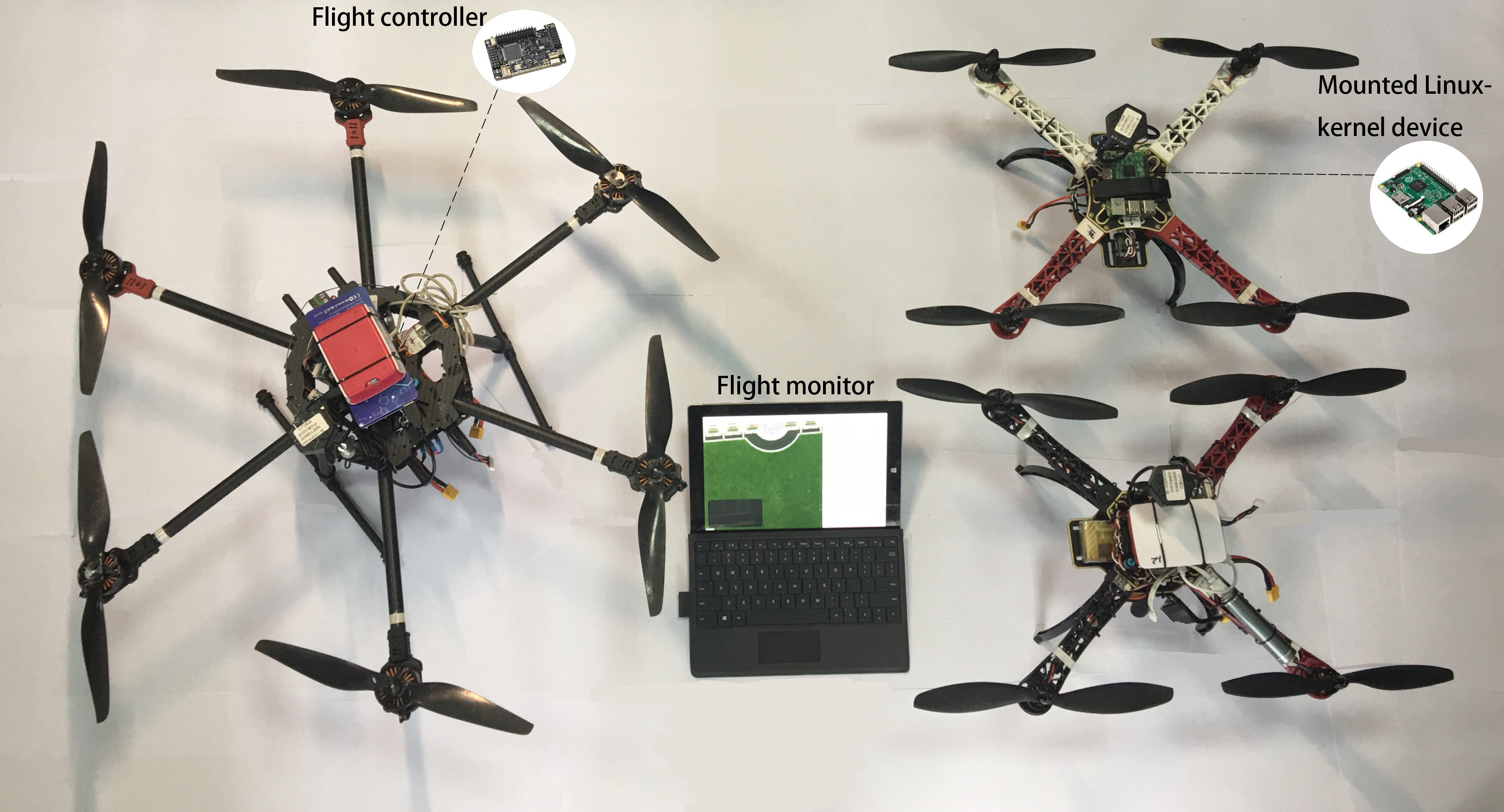}
    \caption{Hardware components in the \systemname platform.}
    \label{hardware}
\end{figure}

\begin{figure*}
    \centering
    \includegraphics[width=.65\textwidth]{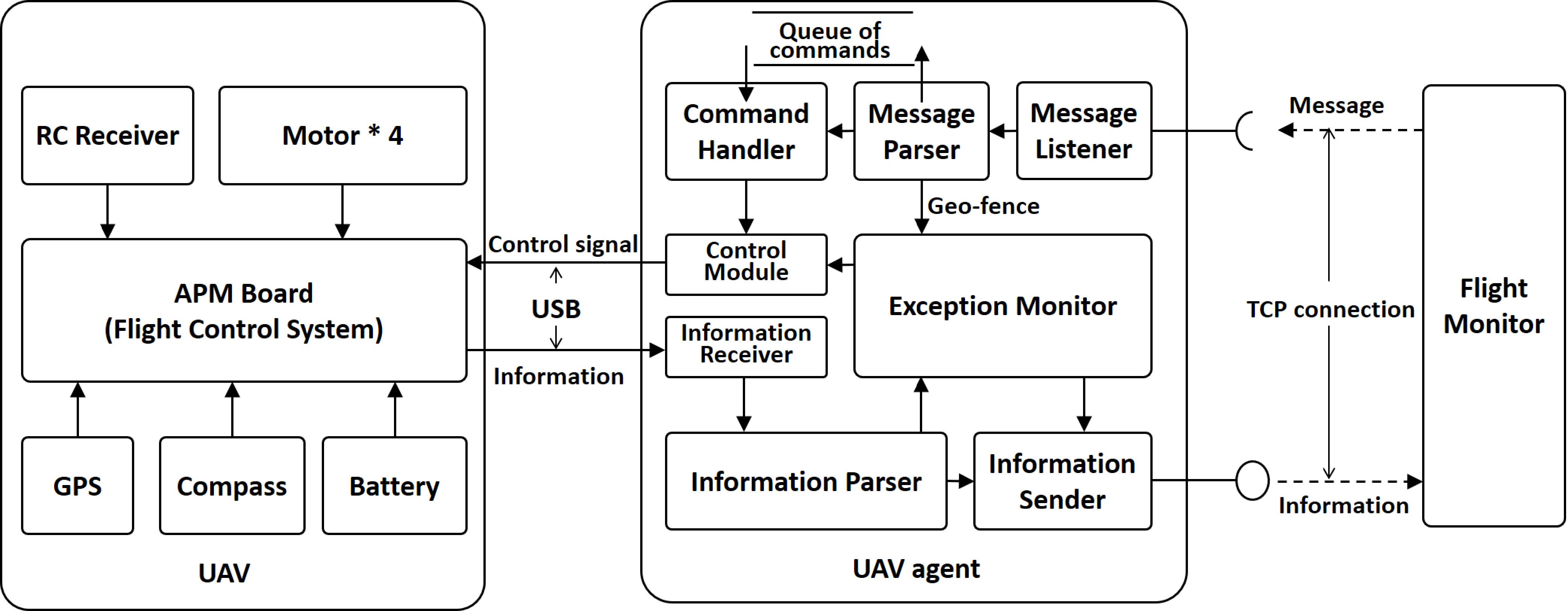}
    \caption{The detailed illustration of the UAV agent and its interaction with the UAV flight controller.}
    \label{fig:uavagent}
\end{figure*}

\subsubsection{Handling status information}

To be compatible to various underlying flight control boards with hardware differences, we access and update the flight status information through SDKs between the flight control module and the UAV agent program. UAV's status information needs to be transmitted to the flight monitor on the ground. Prior to the transmission, the UAV agent periodically parses the flight status (from SDKs) into the formats needed by the exception monitor and the information sender. The information sender further converts it to a character stream for the transmission.

Although the APM2.8 board and Pixhawk HEXA borad are used as the flight control in the current \systemname implementation, our UAV agent design can essentially work with any mainstream UAV flight controllers because the UAV agent program serves as the middleware that hides the UAV difference from the rest parts of the platform, including the flight monitor. Consequently, for UAVs that utilize other flight controllers, our UAV agent can bridge them to the flight monitor, since almost every commodity provider and open-source community offers linux-based SDK for UAV flight control. 


\subsubsection{Handling control messages}
The control message from the flight monitor (on the ground) is transmitted in the format of a character stream, which flows to the message listener of the UAV agent on RPi. There are two types of control messages in \systemname: \textit{control command} and \textit{parameter setting}. For the former type, commands will be appended to a First-In-First-Out queue. The UAV agent has a command handler that can convert each command into the format that is executable by the flight control module. For the latter type, parameters such as geo-fence boundaries, communication range and battery life can be handled by the parameter setting message.

\subsubsection{Monitoring exceptions}

The UAV agent also has an exception monitor  module and the flight status information is periodically sent to this module for inspection. As a result, the exception monitor can track vehicle's status changes and monitor the emergencies. In case any emergency occurs, the exception monitor either delivers high-priority commands to the flight controller or reports to the flight monitor through the information sender. Exceptions in \systemname include low battery, crossing the geo-fence boundary, and bad health of connection to the monitor application, etc.

\subsection{The Flight Monitor}

The main task of the flight monitor is to communicate with each individual UAV and further offers a series of inevitable services for their cooperation. In addition, the flight monitor also provides the interfaces to interact with upper-layer applications and end users through APIs and GUI, respectively. Fig. \ref{fig:gui} demonstrates the GUI from the flight monitor in \systemname platform. 

\begin{figure}
    \centering
    \includegraphics[width = .50\textwidth]{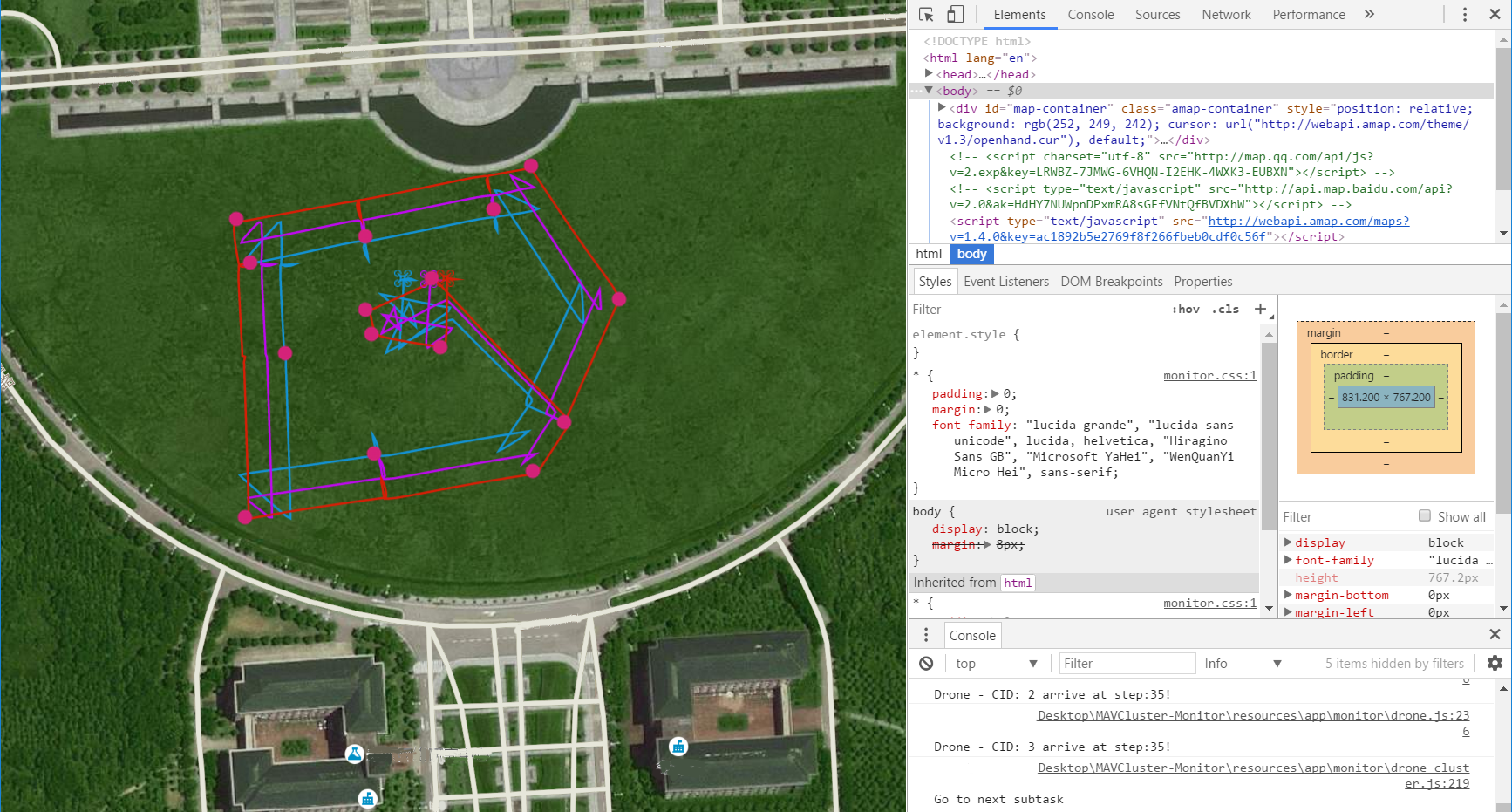}
    \caption{ Example of the  GUI of the flight monitor.} 
    \label{fig:gui}
\end{figure}

\subsubsection{Cooperation services}
\begin{figure*}
    \centering
    \includegraphics[width = .8\textwidth]{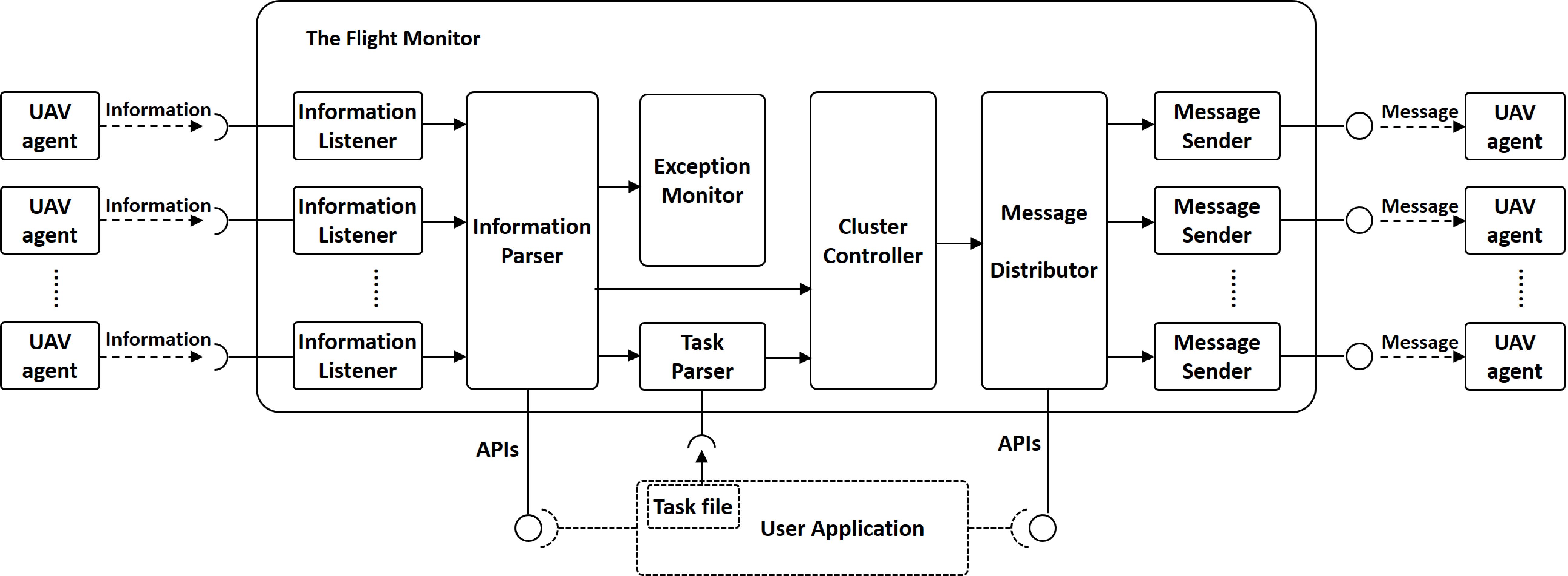}
    \caption{The detailed illustration of the flight monitor and its interaction with UAV agents and applications.}
    \label{fig:my_label2}
\end{figure*}

According to the information received from each UAV, as illustrated in Fig.~\ref{fig:my_label2}, the service controller in the flight monitor could generate control messages to enable the following cooperation services. The control messages are high-priority commands to be sent by the message senders through the message distributor.
\begin{itemize}

\item Connectivity maintenance:  \systemname   enables to connect all UAVs as well as the flight monitor by configuring networking services (Wi-Fi, OLSR) and all the connected UAVs are managed by the agent manager. If the connectivity quality between any UAVs is weak or the transmission errors occur, such exceptions will be thrown and delivered back to the flight monitor to adjust the locations of the UAVs.

\item Path planning: A multi-UAV fleet usually needs to visit a target area collaboratively. The path planning service can well schedule the trajectory of each UAV, so that the fleet can move with connectivity and complete the task from applications at the earliest time. This service will be detailed in Section~\ref{sec:application}.

\item Synchronization: We also need a sequence of synchronization values to indicate the status of each UAV and support the connectivity maintenance and cooperation. Each synchronization value is thus a Boolean type, \eg ``1'' for the collision avoidance of one UAV means this UAV has no potential risk to collide with other UAVs.  

\item Divergence avoidance: Due to the environmental influence (\eg the wind), UAVs may diverge from their planned trajectories or the original location while hovering in the air. By comparing the real-time trajectory with the planned result, the divergence can be captured and further compensated.    

\item Collision avoidance: GPS is widely used to obtain UAV's location. Although GPS is not accurate enough to precisely determine whether two UAVs collided, the collision can be avoided by checking the velocity vectors of any two vehicles and calculating their mutual distances, which should have a sufficient margin for the collision avoidance.
\end{itemize}

\subsubsection{API interaction}
The upper-level user applications can choose to use swarm APIs to implement the aforementioned services. Swarm APIs are implemented mainly by the message center which handles and delivers information and messages bidirectionally.
As shown in Fig~\ref{fig:my_label2}, the Information Parser module handles every status information from drones, and the Message Distributor module is responsible for handling all control messages to the UAVs.
Hence, the status APIs that provide the UAV status information are integrated into the Information Parser and the control APIs that provide the control function and are integrated into Message Distributor.
To satisfy specific requirements on interacting with users of different tasks, GUI APIs that help the developer to quickly create visual interfaces are also provided.   

\subsubsection{Task file interaction}
The upper-level user applications can also choose to interact with the flight monitor through task files in \systemname. 
A task file is an ordered list of actions, while an action consists of a set of key-value pairs. 
There are two types of pairs, \eg compulsory pair and optional pair. 
Compulsory pairs appear in every action, which define the basic of this action.
Examples of compulsory pairs include: basic action and its type, connection ID and its value, synchronization and its boolean value.
Optional pairs are important supplementary to the compulsory pairs, but do not necessarily appear in every action.
Example of optional pairs include: relative distance and its value, absolute destination and its value. 


\subsubsection{UAV communications}
\systemname enables UAVs as well as the flight monitor to be connected by a high performance Wi-Fi   for both the data plane and the control plane. 
Through the Wi-Fi network, TCP connections are set up for data and control transmission. Each packet is composed of the header and the body. To distinguish different types of packets in \systemname, we specify a field in the packet header and define seven    types of packets, as shown in Table~\ref{table_notation}. The packet body contains important information, e.g., actions in a sub-task, center position and radius of the geo-fence, etc. To support more functions in the future, new types of packets can be defined and added. As UAV agents run on linux-kernel devices,    \systemname also supports other networking scheme like the \emph{optimized link state routing protocol} (OLSR) \cite{OLSR}, which is  very common for setting a wireless mesh network. 

\begin{table}
  \centering
  \caption{Types of packets}
 \begin{tabular}{r|l}
  \hline
  Value & Description \\
  \hline
  \hline
  0 & Request of the connection ID \\
  1 & Response to the request of connection ID \\
  2 & Report the status of UAV \\
  3 & Set the geo-fence \\
  4 & Perform action(s) \\
  5 & The synchronization signal \\
  6 & Signal of closing connection \\
  \hline 
\end{tabular}
\label{table_notation}
\end{table}

\subsubsection{Synchronization}
To support the cooperation and smoothly execute the tasks cross UAVs, different UAVs synchronize with each other regularly to cope with asynchronous situations (\eg location deviation, low connectivity quality or different transmission delays). 
To this end, a large task sent to each UAV must be divided into a sequence of small sub-tasks, called steps. 
Each step contains at most one action with synchronization setting true (needs to be synchronized) such that UAVs will be synchronized at the end of each step before going to the next step. 
After its step completed, the UAV agent sends a synchronization message to the flight monitor, and wait until it receives a confirmation message.
After collecting all synchronization messages, the flight monitor sends confirmation to all of the UAVs to continue the next step.
By carefully partitioning the task, the \systemname minimizes the waiting time.

\subsubsection{Energy Simulation}


In the energy simulation, the predicted energy consumption of a flight in the simulation mode must be able to reflect the situation of real flight.  To calibrate the gap between the simulator and the real flight, we train a model by developing a two-step learning framework. The first learning step is to learn a model that maps flying time and distance in real flights to energy consumption. This model is trained through extensive field tests using the non-linear kernel ridge regression. The flying time and distance generated in the simulations might be different from the flying time and distance in real flights. To calibrate such a gap, we further learn another model to map flying time and distance in simulation to flying time and distance in real flights. To this end, we apply a simple linear regression and learn  the parameters that reflect the respective weights/importance of simulated and real flying time/distance.    With such a two-step learning model, we are able to map the flying time and distance in simulations to its expected energy consumption with high accuracy. 
Note that the design of GUI is omitted for space limitation.

\section{Path Planning with Connectivity and Coverage Constraints}
\label{sec:application}

\begin{table}
  \centering
  \caption{Notations}
 \begin{tabular}{r|l}
  \hline
  Symbol & Semantics \\
  \hline
  \hline
  $P$ & set of target points \\
  $p_i$ & $i^{th}$ target point \\
  $x_i$ & horizontal position of $p_i$ \\
  $x_i$ & vertical position of $p_i$ \\
  $U$ &  set of UAVs \\
  $u_i$ & $i^{th}$ UAV \\
  $u_{j,t}$ & position of $u_j$ at time $t$ \\ 
  $x_{j,t}$ & horizontal position of $u_j$ at time $t$ \\
  $y_{j,t}$ & vertical position of $u_j$ at time $t$ \\
  $d$ & distance a UAV can move during a flight \\
  $\omega$ & UAV's transimission range \\
  $x_{i,j,t}$ & whether the distance between $u_i$ and $u_j$ is less than $\omega$ at time $t$ \\
  $y_{i,j,t}$ & whether target point $p_i$ is scanned by $u_j$ at time $t$ \\
  $L_j$ & total moving distance of $u_j$ \\
  \hline 
\end{tabular}
\label{table_notation_definition}
\end{table}

To verify the \systemname platform functionality, we detail an emergency search application in this section to provide an embedded path planning service for the multi-UAV cooperation by considering both the UAV network connectivity and coverage.
We will first introduce the problem, then describe an algorithm for the cooperative path planning briefly for space limitation.

\subsection{Target-point Visiting Problem Formulation}
Assume in a two-dimensional area,  a set of target points need to be visited/covered by any UAV of a fleet.
Let the point set be $P=\left \{ p_1, p_2, \ldots, p_n \right \}$, where $p_i = (x_i, y_i)$ refers to the position of the $i$-th target point. 
Let $U = \left\{u_1, u_2, \ldots, u_m \right\}$ be a set of $m$ UAVs. We assume that these UAVs can move freely and we do not consider the vertical movement.
At time slot $t$, the position of $u_j$ is denoted by $u_{j,t}=\left(x_{j,t},y_{j,t} \right), t \in [0,T)$. 
If one UAV $u_j$ passes through target point $p_i$, we say that $p_i$ is covered (or scanned) by $u_j$. The notations are summarised in Table \ref{table_notation_definition}. Note we assume all UAVs move at a constant speed in the planning stage. If a UAV's velocity changes due to wind disturbance during a flight  journey, such a change can be captured by the monitoring module and replanning will be triggered.

Since the flight speed of the UAV is limited, the distance that a UAV can move during a time unit should satisfy the \textit{speed constraint},
\begin{equation} \label{eq:speed-con}
d( u_{j,t}, u_{j,t+1} ) \leq d_{max}, \quad \forall u_j \in U, \ \forall t \in [0,T)
\end{equation}
where $d_{max}$ is the largest distance that a UAV can move within a time slot.

During a flight, a UAV needs to keep connected with at least one UAV in the fleet or the ground controller to ensure the information captured by the UAV can be quickly transmitted back to the ground controller and UAVs can cooperatively complete the task. This constraint is referred to as \textit{connectivity constraint}.   In other words, the inter-UAV distance shall be within the UAVs' transmission range, denoted by $w$.  We use $x_{i,j,t}$ to denote whether the distance between $u_i$ and $u_j$ ($i \ne j$) is less than $w$ at time $t$, that is, 
\begin{equation}
x_{i,j,t} = 
\begin{cases}
1, & \text{if } d(u_{i,t}, u_{j,t}) \le w \\
0, & \text{if } d(u_{i,t}, u_{j,t}) > w
\end{cases}
\end{equation}
Then the connectivity constraint can be formulated by the following inequality: 
\begin{equation} \label{eq:connect-con}
\sum_{j: u_j \in U, \ j \ne i} d(u_{i,t}, u_{j,t}) \cdot x_{i,j,t} > 0, \quad \forall u_i \in U, \ \forall t \in [0, T)
\end{equation}

Each target point must be scanned by at least one UAV, we call it the \textit{coverage constraint}. We use $y_{i,j,t}$ to represent whether target point $p_i$ is scanned by $u_j$ at time $t$, that is, 
\begin{equation} \label{3}
y_{i,j,t}=\left\{
\begin{aligned}
1, & & \text{if } u_{j,t} = p_i \\
0, & & \text{otherwise}
\end{aligned}
 \right.
\end{equation}
Then the coverage constraint can be denoted by 
\begin{equation} \label{eq:coverage-con}
\sum_{j: u_j \in U}\sum_{t: t \in [0,T)} y_{i,j,t} \geq 1, \quad \forall p_i \in P
\end{equation}

For emergency search applications, it is critical to complete the journey as soon as possible. On the other hand, considering limited battery supply on UAVs, it is also critical to minimize the longest route that a UAV flies. Thus, we define our objective as
\begin{equation} \label{eq:cost}
L_{fleet} = \max_{u_j \in U} L_j
\end{equation}
where $L_j$ is the total moving distance of $u_j$, it can be calculated by $L_j = \sum_{t=0}^{T-1}d(u_{j,t}, u_{j,t+1})$. 

We formally define the multi-UAV target-point visiting problem as follows.
\begin{definition} \label{def:problem}
Given a fixed number of UAVs and a set of target points distributed in an area, the multi-UAV target-point visiting problem is to plan the trajectory that each drone should move along at each time, to minimize the longest route, 
\begin{alignat*}{2} \label{eq:opt_target}
 \min\max_{u_j \in U} L_j\ \ \ \ \ \ \ \ \\
\ \ \ \ \ \ \ subject\ to\ Eq.(\ref{eq:speed-con})(\ref{eq:connect-con})(\ref{eq:coverage-con})
\end{alignat*}
\end{definition}

\subsection{ Polygon-Guided Scanning Algorithm} \label{sec:CSA}
In this section, we propose a UAV fleet planning algorithm, called \textit{Polygon-guided Scanning Algorithm} (PSA),  to cooperatively visit the target points as soon as possible. We note that no matter what kind of planning method is applied, the set of points that constitute the convex hull of all the target points should be scanned. Thus, our high level idea is to let the UAV fleet scan the area   by first moving along the convex hull of the target points and cover the maximum possible range under communication connectivity constraint; Since there may exist target points that cannot be touched  in the first round of scanning, the fleet will iteratively scan the remaining target points round by round. Such a scanning process would partition the area into sub-areas  through iterations. Observing that the optimal solution needs  to incur   a travel distance that is at least  the  perimeter of the convex polygon, we will design a partition method that ensures  the scanned area in each iteration/round is the maximum possible, making the travel distance of the fleet approach that of the optimal solution.  

In the following subsubsections, we will first introduce the partition of the area and the overall scanning scheme for the UAV fleet, then we will introduce the detailed scheme for a  round of scanning. Finally we will give the performance analysis of the proposed algorithm.

\subsubsection{Overall scanning scheme} \label{subsec:overall}

In this subsection, we introduce  the overall scanning scheme on how  to divide the area into sub-areas. Given  all the current non-scanned target points, we can compute the corresponding convex hull, the minimum convex set that contains all the target points to be scanned. Let $V_k = \{v_{k,1}, v_{k,2}, \ldots, v_{k,h_k} \}$ be the computed convex hull in the $k$-th round of scanning, where  $h_k$ is the number of vertices of the convex hull. 
The convex hull $V_k$ is called the \textit{outer boundary}.  We make UAV fleet   move along the outer boundary  and cover the maximum possible range  under the communication connectivity constraint (keep the formation to be evenly spaced on a line with a length   $w_{max} = (m-1) \times w$). This would generate  a ring-like area, denoted by $R_k$,  as illustrated in Fig. \ref{fig:2}(a), which is exactly the area of the $k$-th round of scanning.
Let $V_k' = \{v_{k,1}', v_{k,2}', \ldots, v_{k,h_k}'\}$ be the points constituting the inner boundary of the ring $R_k$. Obviously, the inner bounder  is parallel to and lies inside the outer boundary. 
Let $P_k$ be the set of target points inside the area $R_k$ (e.g., for $k=1$, $P_1 = P \cap R_1$). The specific scanning scheme for target points $P_k$ will be described in detail later.

After scanning the   set $P_k$, the remaining target points $P \setminus \bigcup_{i= 1}^{k}P_{i}$ all locate inside the inner boundary of $R_k$. We continue the scanning by treating $P \setminus \bigcup_{i= 1}^{k}P_{i}$ as new given target points and computing the convex hull of these points again to determine the scanning area $R_{k+1}$ in the next round of scanning. We repeat this operation until all the target points are scanned. Let the total number of rounds be $K$, then $P = \bigcup_{k=1}^{K}P_k$. Fig. \ref{fig:2}(a) shows a schematic diagram when $K = 3$. The detailed description of the overall scanning scheme is shown in Algorithm \ref{alg:PSA}. 

\subsubsection{Transfer scheme  of two adjacent rounds} Now we introduce the transfer scheme on how the UAV fleet will travel from a scanned area $R_k$ to the next area $R_{k+1}$ to be scanned in the next round, as shown in Fig. \ref{fig:2}(b). In our scheme, the UAV fleet will travel from $R_k$ to $R_{k+1}$ along a line as follow. 
During the process that all the drones move from the outermost scanning area $R_1$ to the innermost one $R_K$, the shortest distance between the outer boundary of $R_1$ and the outer boundary of $R_K$ is treated as the shortest \textit{transfer distance}. It is not difficult to see that this value is the shortest distance from the points in $V_K$ to the outer boundary of $R_1$.
Let the intersection points between the line with the shortest transfer distance and the outer boundary of $R_k, k=1,2,\ldots,K$ be $C = \{ c_{1}, c_{2}, \ldots, c_{K} \}$.   Fig. \ref{fig:2}(b) illustrates a segment of $c_{1} c_{K}$ that achieves the shortest transfer distance, which is denoted as $L_{trans}$. 
In the $k$-th round of scanning, the fleet start the flight from $c_k$, after scanning the area $R_k$, the fleet return to $c_k$, and then move along $c_{k} c_{k+1}$ to $c_{k+1}$ and then start the next round of scanning.
\begin{remark}
Although the fleet needs to adjust the formation when all the drones enter the next scanning area, considering the length of the flight along the outside of a scanning area is usually much longer than that along the inside, the moving distance contributed by this adjustment can be ignored, which implies that under our planning the longest route of the UAV fleet is determined by the  drone which moves along the outer boundary of each area $R_i, i=1,2,\ldots,K$.
\end{remark}

\begin{figure*}
\centering
\includegraphics[width=0.65\textwidth]{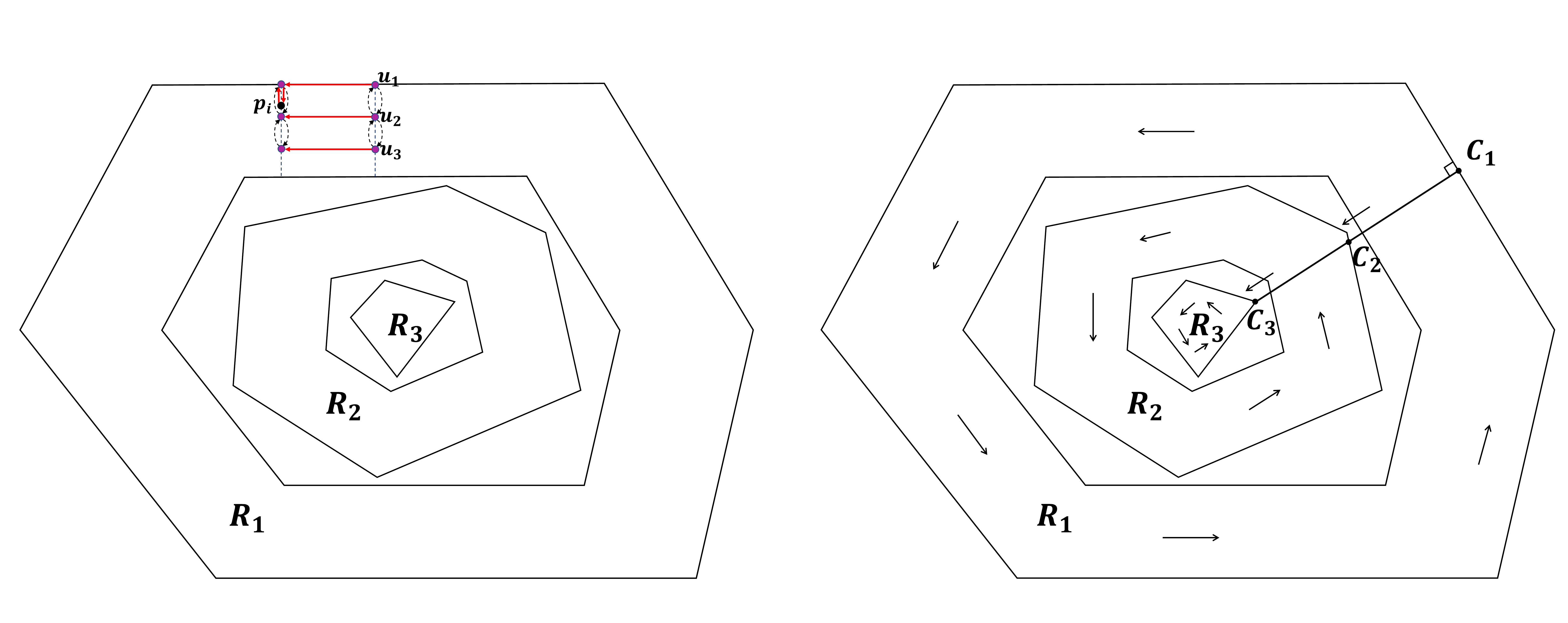} 
 \caption{ (a) The partition of the area where $K=3$; (b) The transfer of UAV fleet from one scanning area to the next scanning area when $K=3$ }
 \label{fig:2}
\end{figure*}

\begin{algorithm}[t]
\caption{Polygon-guided Scanning Algorithm (a.k.a., PSA)}  \small
\hspace*{0.02in} {\bf Input:} 
the set of target points $P$, the set of drones $U$, maximum communication distance $w$\\
\hspace*{0.02in} {\bf Output:} 
the flight planning and the corresponding flight cost of the UAV fleet $L_{fleet}$

\begin{algorithmic}[1]
\State Let the largest scanning width be $w_{max} \leftarrow (m-1) \times w$, let $k \leftarrow 1$, let $L_{fleet} \leftarrow 0$;
\While{$P$ is not empty} 
　　\State Compute the set of points $V_k$ that constitute the convex hull of $P$ (e.g., by adopting the method of Graham scanning); \label{alg-psa:scanning-area-s}
　　\State Compute the set of points $V_k^{'}$ that constitute the inner boundary of the area of the current round of scanning, based on $V_k$ and $w_{max}$; // by making lines parallel to $v_{k,h} v_{k,h+1}$
　　\State Get the sub-area $R_k$ and target points $P_k$ in round $k$ based on $V_k$, $V_k^{'}$ and $P$; \label{alg-psa:scanning-area-e}
　　\State Call Algorithm \ref{alg:psta} to scan sub-area $R_k$ and get the corresponding flight cost $L_k$; \label{alg-psa:psta}
　　\State $L_{fleet} \leftarrow L_{fleet} + L_{k}$;
　　\State $P \leftarrow P \setminus P_k$;
　　\State $k \leftarrow k + 1$;
\EndWhile
\State Compute the transfer distance $L_{trans}$;
\State $L_{fleet} = L_{fleet} + L_{trans}$;
\State \Return $L_{fleet}$
\end{algorithmic}
\label{alg:PSA}
\end{algorithm}

\subsubsection{The planning for single round of scanning} \label{subsec:single}
Now we introduce the detailed planning for scanning target points in a single round, the specific ring-like area $R_k$.

As mentioned before, during the process of scanning target points, we will keep all the drones on a line. We call these lines formed by the UAV fleet the \textit{scanning lines}. We will make the scanning lines \textit{perpendicular} to the parallel lines of the outer and inner boundary of area $R_k$, at any time. The scanning process then can be viewed as the traversal of the scanning lines. Fig. \ref{fig:scheme-and-adjustment}(a) illustrates the scanning lines (dashed lines) during the scanning. In general, UAVs in the fleet formation will travel in parallel along the outer boundary, except the case that some target point is covered by the scanning line of the fleet during the flight.   In that case, the drones should adjust and travel along the scanning line to cover  the target points that fall on this line. Besides, in our planning, the distance between any two adjacent drones remains unchanged all the time, except for the case when they meet a vertex of the scanning area where the drones need to adjust their directions.

We first determine the flight cost caused by traversing between the scanning lines. Assume that the perimeter of the outer boundary (convex polygon) in area $R_k$ is $L_k$. As shown in Fig. \ref{fig:scheme-and-adjustment}(a), during the traversing between the scanning lines in the area $R_k$, the outermost drone $u_{1}$ travels along the convex polygon and  contributes the maximum flight distance. Thus, the flight cost of the fleet caused by traversing between the scanning lines is $L_k$. 
\begin{figure}
    \includegraphics[width = 1.0\columnwidth]{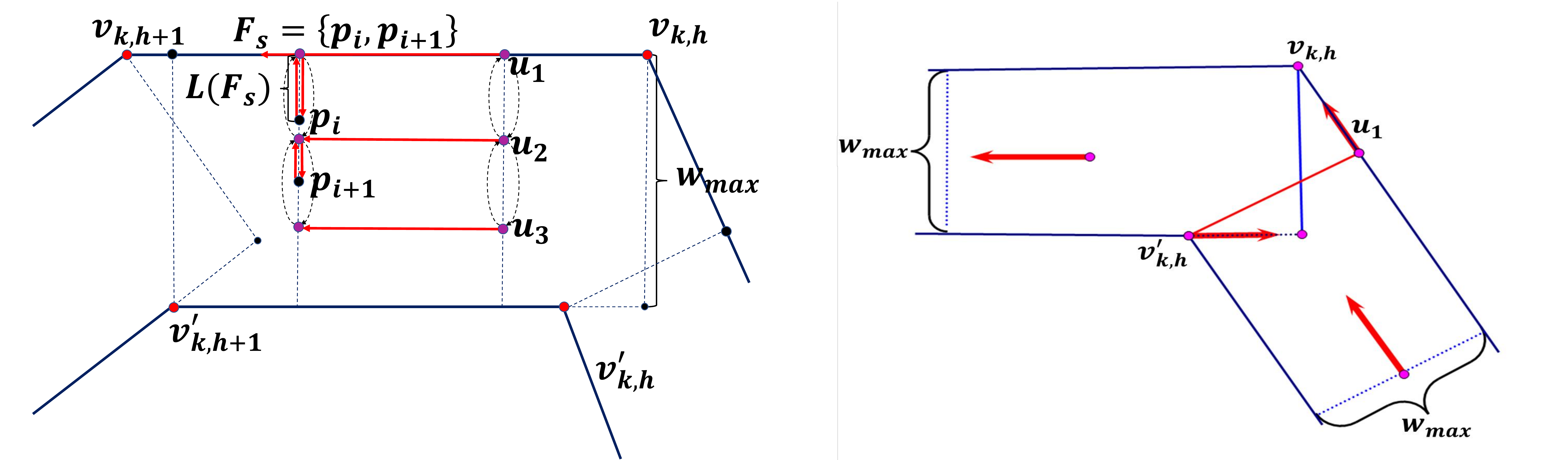}
    \caption{(a) The scheme for single round of scanning; (b) Adjustment in the corner region. }
    \label{fig:scheme-and-adjustment}
\end{figure}

Then, we determine the  flight cost to adjust the fleet and travel along the scanning line for visiting target points lying on a single scanning line. 
Since   target points must be passed by   scanning line, we can divide the set of target points $P_{k}$ in $R_k$ into a number of subsets. Denoted by $F_s$ the $s$-th subset that contains  target points lying on the same   scanning line. Fig. \ref{fig:scheme-and-adjustment}(a) illustrate a set $F_s$ with two target points that lie on the same scanning line.  Assume that there are  $S$ such subsets. Then, $P_{k} = \bigcup_{s= 1}^{S}F_{s}$ where $ F_{i} \cap F_{j} = \varnothing (i \ne j)$.   

\begin{algorithm}[t]
\caption{Polygon-guided Scanning Single-Turn Algorithm (PSTA)} \small
\hspace*{0.02in} {\bf Input:} 
the set of drones $U$, the scanning area $R_k$, the set of target points $P_{k}$, maximum communication distance $w$ \\
\hspace*{0.02in} {\bf Output:} 
the flight planning and corresponding flight cost of the UAV fleet $L_{k}$ for the $k$-th round of scanning 

\begin{algorithmic}[1]
\State Compute the subset of points $F_{s} (s = 1,2,\ldots,S)$ where each subset includes all the target points lying on the same perpendicular line of the two boundaries; \label{alg-csta:divide}
\State Compute the length $L(b_k)$ of the outer boundary of $R_k$; \label{alg-csta:lbk}
\State Let $L_k \leftarrow L(b_k)$;
\For{$s=1$ to $S$} \label{alg-csta:scan-s}
    \State Let the UAV fleet move to the scanning line which passes through $F_s$;
　　\State Let the fleet fly towards the inner boundary along the scanning line, until all the target points in $F_s$ are scanned, and then return back.
　　\State Let the flight cost caused by travelling along the scanning line be $L(F_s)$;
　　\State $L_k \leftarrow L_k + 2 \cdot L(F_s)$; \label{alg-csta:scan-e}
\EndFor
\State \Return $L_k$
\end{algorithmic}
\label{alg:psta}
\end{algorithm}

In our planning, in order to obey the connectivity constraints of all the drones, the whole fleet will make an adjustment and fly towards the inner boundary of $R_k$ simultaneously to touch the target points covered by the current scanning line. If there are multiple target points on the scanning line, the flight cost will be determined by the minimum moving distance to scan all the target points, which can be easily measured. Let  $L(F_s)$ be such a cost generated on a scanning line $F_s$.  When all the target points on the current scanning line have been scanned, the fleet will return to their starting points on this scanning line and move on to the next scanning line, as demonstrated by the red lines in Fig. \ref{fig:scheme-and-adjustment}(a). Therefore, the flight cost cause by the adjustment would be $2 \cdot L( F_{s})$. We use  $L_{adjust}=\sum_{F_s\subseteq  P }2L(F_s)$ to denote the  total travel distance generated by the minor adjustments during the whole flight.  

Note that  after finishing scanning along an edge  of the outer boundary, we need to further adjust the fleet formation and make it  scan in parallel along the next edge  of the outer boundary, as illustrated in Fig. \ref{fig:scheme-and-adjustment}(b). Such an adjustment would not increase the flight cost of the fleet, since although it causes extra travel distance of other UAVs (except the outermost one),  the outermost UAV still travels with the longest distance  along the outer boundary.   

Algorithm \ref{alg:psta} presents the details of the scanning process for a specific sub-area. Line \ref{alg-csta:divide} constructs the subsets $F_s$. Line \ref{alg-csta:lbk} computes the total flight cost when the fleet traverses between scanning lines. Line \ref{alg-csta:scan-s}-\ref{alg-csta:scan-e} guide the fleet to scan the target points  $F_s$ falling on a scanning line and computes the corresponding flight cost.

\subsubsection{Theoretical analysis}
We further theoretically analyze the worst-case performance of the algorithm. In intuition, according to the partition scheme of PSA, the convex polygon generated in each round of PSA is the minimum convex set containing the remaining non-visited target points, thus the optimal solution needs to travel with a distance equaling the perimeter of the convex polygon in each round. This allows us to bound the difference of UAV fleet travel  distance between PSA and the optimal solution. Let $L(PSA)$ and $L(OPT)$ be the UAV fleet travel distance incurred by algorithm PSA and the optimal solution, respectively. Note that during the traveling along convex polygons, or more exactly  traversing between scanning lines, the UAV fleet needs to adjust and move along the scanning line to cover the target points on the scanning line.     In the following theorem, we show that the extra travel distance generated by   PSA compared to the optimal solution is upper bounded by $L_{adjust}$, which is usually a short moving distance caused by minor adjustment when traversing along the convex polygons.  

\begin{theorem} \label{theorem:performance}
The gap of the flight distance between PSA and the optimum solution satisfies \(L\left ( PSA \right )-L\left ( OPT \right )\leq L_{adjust}\).  
\end{theorem}

\begin{figure*}
    \centering
    \includegraphics[width = .6\textwidth]{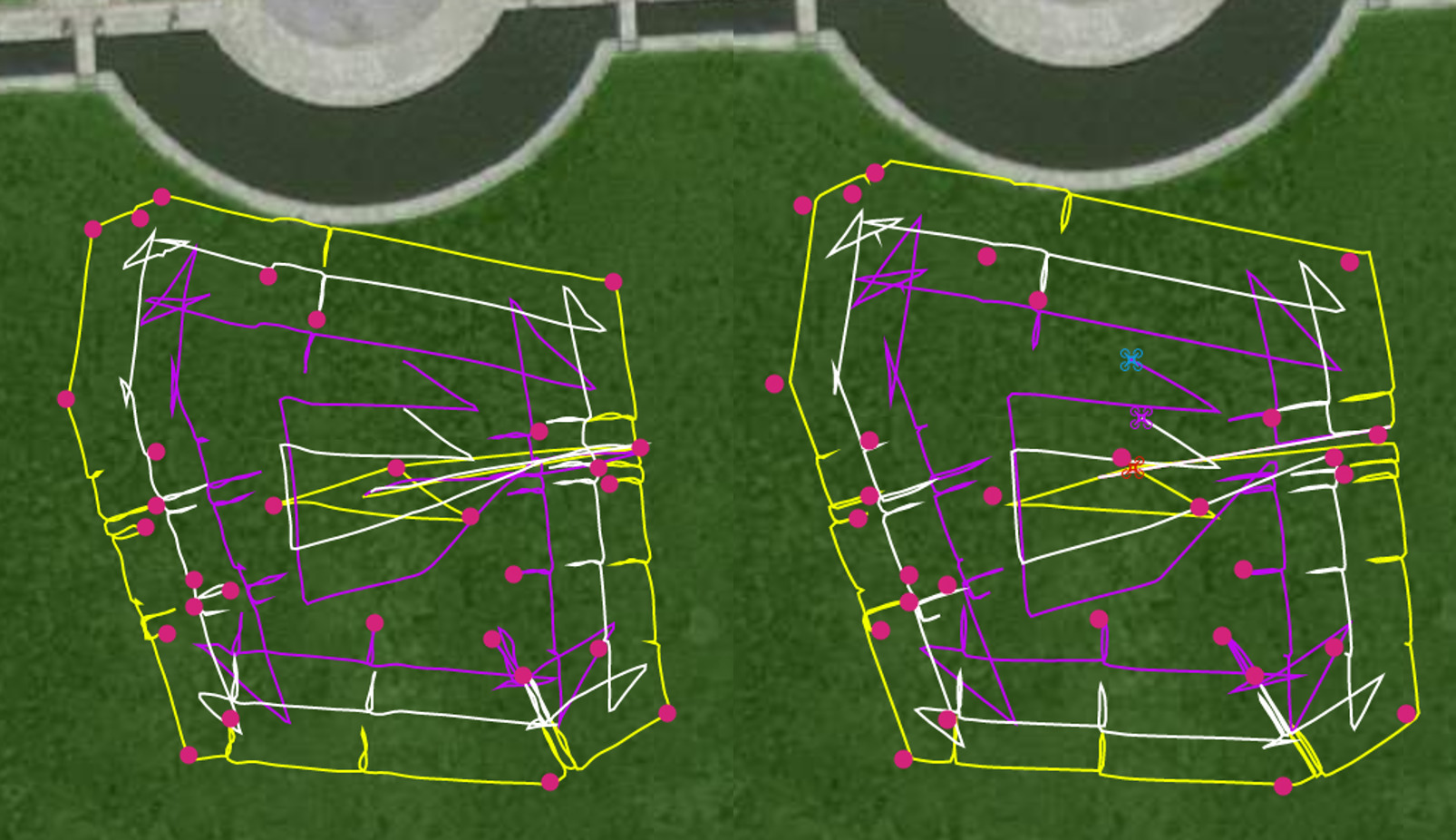}
    \caption{ Trajectories of 3 UAVs   in field test (left) and simulation environment (right) when completing a task with 30 points.}
    \label{fig:test1}
\end{figure*}

\begin{proof}
  There are three constraints  in the target-point visiting problem. The coverage constraint requires us to scan all the target points in target points set $P$, which indicates that we must scan the vertices on the convex hull of target points. On the other hand, connectivity constraint requires the connectivity of drones in real time, hence the largest scanning width of drones is \(w_{max}\). According to the partition scheme of PSA,    each  sub-area \(R_{k}\) is the maximum   region  that the UAV fleet can scan along the convex polygon. Therefore, in the optimum solution, the  distance   the UAV fleet needs to travel within each sub-area \(R_{k}\) is at least \(L_{k}\). Moreover, as \(L_{trans}\) is the shortest distance from the outermost boundary to the innermost boundary, the optimal solution should also incur another travel distance \(L_{trans}\) to move the fleet from the outermost area to the innermost area. Hence, we can get the lower bound of the optimum solution that \(L\left ( OPT \right )\ge  L_{trans}+\sum_{k= 1}^{K}L_{k}\).
  
  Noting that the travel distance of the UAV fleet generated by algorithm PSA contains three parts, the distance $L_{k}$ incurred by travelling along the convex polygon (which is the distance the outermost UAV travel with), the distance $L_{trans}$  incurred  by moving along the transfer line, the distance $2L(F_s)$ incurred by covering the target points on a scanning line $F_s$. Accordingly, the total travel distance of PSA is $\sum_{k=1}^K L_k+L_{trans}+L_{adjust}$ where $L_{adjust}=\sum_{F_s\subseteq P}2L(F_{s})$. Therefore, the extra travel distance generated by PSA compared to the optimal solution is at most $L_{adjust}$.
\end{proof}

\subsection{Implementation  as path planning service }
The planned paths can be   converted     to a series of control commands and disseminated   to individual UAVs for real flight by leveraging the  core cooperation service  of \systemname (synchronization, connectivity maintainence, divergence  avoidance, etc.). 

Based on the planned path information, the energy simulation module of the simulator would predict the energy to be consumed under the current configuration (e.g., number of UAVs, battery sizes). In case energy/batteries are not feasible to support the flight, users can upgrade the configuration (e.g., enlarge the number of UAVs or battery sizes) until it admits a feasible flight. 

 Note that the path planning service provided by \systemname can be further configured and adaptive to  other upper-level user applications by leveraging the  core cooperation service if necessary.

\section{Experimental Results and Analysis}
\label{sec:results}

In this section, we perform field tests and simulations to demonstrate the effectiveness of our \systemname platform in  simulating UAVs cooperation and real-world flights. We will perform the tasks in both the simulator environment and the field tests,   demonstrating that our simulator can be used as a reliable platform for testing.

\subsubsection{Comparison between the trajectories generated in the simulator and field test for a single task}
We first set a target-point visiting task that consists of   30 target points and 3 UAVs as a fleet. The length and width of the field are set to be 120m. All target points are randomly generated in this area. By default, the tasks are performed  at the speed of 4m/s.   We first run the task in our simulator to generate trajectories and predict the energy consumption based on the energy  model trained for UAVs. Then, we use the following two types of UAVs  in our field tests, copters with frame QUAD or HEXA controlled by Pixhawk, copters with frame QUAD controlled by APM.    The distance of a task is calculated according to the real-time coordinates from GPS. The time of flight is calculated according to the system clock of Raspberry Pi 3b on UAV. The energy consumption is measured by computing the real-time voltage and current of battery read from the power module.

 Fig. \ref{fig:test1} shows the trajectories generated in our \systemname simulator (left one) and the field test (right one). From the figure, we can see that the trajectories generated in the simulator is almost the same to that in the field tests.  The trajectory is more irregular in field test between two inflection points, which is caused by the unstable environment conditions in reality. Despite of this, after examining log data from the connectivity management and synchronization modules of \systemname, we find that the fleet manages to maintain the connectivity and visits all target points while satisfying the battery capacity.

 \subsubsection{Comparison of flying distance and energy consumption between simulation and field test for a single task}

For the same task, we recorded the flying distance and energy consumption during the whole flight in both simulation and the field test.  Fig.~\ref{fig:test-flight} shows the flying distance and energy consumption on one of the three UAVs, which is a copter   with  frame   HEXA  controlled by  Pixhawk. We can see from the figure that during the whole flight, at any given time, the flying distance and consumed energy in simulation and field test is consistently close. This demonstrates that simulator can reliably reflect the real flight. 

\begin{figure}[tb]
\centering
\subfigure[Flight distance versus flying time.]{
\begin{tikzpicture}[scale = 0.48]
\pgfplotsset{every axis legend/.append style={
at={(0.5,1.03)},
},every axis y label/.append style={at={(0.07,0.5)}}}
\begin{axis}[xlabel=Flying time (s),
    ylabel=Flight distance (m),xtick ={0,100,200,300,400,500,600},legend style={{font=\footnotesize},at={(0.01,0.98)},
anchor=north west,legend columns=2}]
\addplot [draw=red] [x=a, y=Field test of 2 drones] 
coordinates{
(0,0)
( 19.609237999999998, 35.28420612954723)
( 38.999651, 91.01741891586072)
( 58.419162, 114.0433985746312)
( 77.82266299999999, 163.41521396606467)
( 97.212115, 231.71578832993598)
( 116.617852, 256.4169573127502)
( 136.030384, 296.7099695124614)
( 155.41580199999999, 319.0456034646929)
( 174.82853699999998, 368.0711190865987)
( 194.227072, 397.2620254329996)
( 213.631203, 422.23440248033546)
( 233.037399, 449.82454722592865)
( 252.43917499999998, 473.7220322233372)
( 271.838062, 487.7486757355393)
( 291.239901, 523.6553451715569)
( 310.632347, 565.0318536130289)
( 330.036387, 586.1368793381265)
( 349.444117, 626.5875876318639)
( 368.839434, 675.3230940043103)
( 388.237056, 703.9356582407293)
( 407.65527, 731.0186880849061)
( 427.049106, 774.1249079888931)
( 446.45871999999997, 801.7721138504759)
( 465.860297, 833.4955029798414)
( 485.24758399999996, 864.5347655606898)
( 504.65653299999997, 892.1056023613885)
( 524.061369, 938.6806368837376)
( 543.469199, 995.5667398652895)
( 562.8703909999999, 1054.9970699680393)
( 579.4821509999999, 1061.539792820901)
};
\addplot [draw=blue] [x=a, y=Simulator of 2 drones] 
coordinates{
 (0,0)
( 20.035, 34.077465206771876)
( 40.072, 76.2974846202103)
( 60.104, 99.72508084141917)
( 80.128, 146.66753003138965)
( 100.16, 201.6519185470638)
( 120.199, 226.71891290359642)
( 140.229, 258.7798012509676)
( 160.258, 278.8553353699067)
( 180.291, 313.6382721991811)
( 200.322, 349.7624118161912)
( 220.364, 374.58517203657135)
( 240.393, 398.8519968511438)
( 260.42, 423.76046943434966)
( 280.455, 436.834295934539)
( 300.479, 457.50469254344205)
( 320.51, 497.97353279903297)
( 340.54, 525.2439845992096)
( 360.577, 562.5889999434444)
( 380.603, 601.8174704858686)
( 400.63800000000003, 627.3957902134653)
( 420.657, 652.2469790363268)
( 440.68, 686.6569623001619)
( 460.709, 722.7230682594593)
( 480.743, 746.478249576319)
( 500.776, 773.8031390948627)
( 520.812, 794.3928552684612)
( 540.844, 829.919368681299)
( 560.873, 875.3954816170739)
( 580.904, 932.9374459452091)
( 600.933, 953.1932482636834)
( 608.44, 953.3276574506438)
    };

\legend{Field test, Simulator}
\end{axis}
\end{tikzpicture}
}
~~~~
\subfigure[Energy consumption versus flying time.]{
\begin{tikzpicture}[scale = 0.48]
\pgfplotsset{every axis legend/.append style={
at={(0.5,1.03)},
anchor=south},
every axis y label/.append style={at={(0.07,0.5)}}}
\begin{axis}[xlabel=Flying time (s),
    ylabel=Consumed energy (kJ),xtick ={0,100,200,300,400,500,600},legend style={{font=\footnotesize},at={(0.01,0.98)},
anchor=north west,legend columns=2}]
\addplot [draw=red] [x=a, y=Field test of 2 drones] 
coordinates{
(0,0.0)
( 19.609237999999998,6.214245384568442)
( 38.999651,14.388310790102762)
( 58.419162,22.78070113967443)
( 77.82266299999999,31.01476873038377)
( 97.212115,39.33362055085648)
( 116.617852,47.73682089993064)
( 136.030384,55.99741354865286)
( 155.41580199999999,64.30404294080549)
( 174.82853699999998,72.5837496735891)
( 194.227072,80.89060908950123)
( 213.631203,89.24334328570525)
( 233.037399,97.68206446071561)
( 252.43917499999998,106.03857662162069)
( 271.838062,114.43035690840992)
( 291.239901,122.81645331895488)
( 310.632347,131.1895271327453)
( 330.036387,139.56024398233802)
( 349.444117,147.8560838703903)
( 368.839434,156.06513277996658)
( 388.237056,164.48270912443147)
( 407.65527,172.93212446437406)
( 427.049106,181.1900915662872)
( 446.45871999999997,189.52583267599576)
( 465.860297,197.8857165581287)
( 485.24758399999996,206.22867181646873)
( 504.65653299999997,214.63485189077338)
( 524.061369,222.95335616337786)
( 543.469199,231.1442589582193)
( 562.8703909999999,239.28335332246675)
( 579.4821509999999,245.69571000529967)
};
\addplot [draw=blue] [x=a, y=Simulator of 2 drones] 
coordinates{
(0,0.0010918708396382081)
( 20.035,6.53416325378418)
( 40.072,13.156470840454102)
( 60.104,19.868837677001952)
( 80.128,26.673314331054687)
( 100.16,33.578499267578124)
( 120.199,40.587310791015625)
( 140.229,47.69730029296875)
( 160.258,54.9143916015625)
( 180.291,62.2434794921875)
( 200.322,69.68553515625)
( 220.364,77.2488203125)
( 240.393,84.927263671875)
( 260.42,92.7280859375)
( 280.455,100.65870703125)
( 300.479,108.71429296875)
( 320.51,116.9048828125)
( 340.54,125.2311953125)
( 360.577,133.699953125)
( 380.603,142.3049375)
( 400.638,151.059875)
( 420.657,159.955953125)
( 440.68,169.006265625)
( 460.709,178.214375)
( 480.743,187.57996875)
( 500.776,197.1080625)
( 520.812,206.8013125)
( 540.844,216.6600625)
( 560.873,226.68928125)
( 580.904,236.8894375)
( 600.933,247.26721875)
( 608.44,251.207)
};

\legend{Field test, Simulator}
\end{axis}
\end{tikzpicture}
}
\caption{ Flying distance and energy consumption in simulation and field test for a single task }\label{fig:test-flight}
\end{figure}

\subsubsection{Comparison of flying distance, flying time, energy consumption in simulation and field test over multiple tasks}

\begin{figure}[tb]
\centering
\subfigure[Flight distance for different number of points.]{
\begin{tikzpicture}[scale = 0.43]
\pgfplotsset{every axis legend/.append style={
at={(0.5,1.03)},
},every axis y label/.append style={at={(0.07,0.5)}}}
\begin{axis}[xlabel=Number of points,
    ylabel=Flight distance (m),ymax =1200,xtick =data,legend style={{font=\footnotesize},at={(0.01,0.98)},
anchor=north west,legend columns=2}]
\addplot+ [x=a, y=Field test of 2 drones] 
coordinates{(5,469.438578)  (10,568.87614) (15,765.3315) (20,846.9956) (25,932.847) (30,992.1754)};
\addplot+ [x=a, y=Simulator of 2 drones] 
coordinates{(5,420.905)
    (10,524.868)
    (15,701.013)
    (20,752.411)
    (25,854.104)
    (30,904.386)};
\addplot+ [x=a, y=Field test of 3 drones] 
coordinates{(5,468.1030962)  ((10,580.3055813)
    (15,668.2471601)
    (20,736.454225)
    (25,897.6784925)
    (30,960.1303403)};
\addplot+ [x=a, y=Simulator of 2 drones] 
coordinates{(5,423.51253)
    (10,543.48241)
    (15,611.30578)
    (20,661.71816)
    (25,826.361679)
    (30,870.767576)};
\legend{Field test of 2 drones, Simulator of 2 drones,Field test of 3 drones, Simulator of 3 drones}
\end{axis}
\end{tikzpicture}
}
~~~~
\subfigure[Energy consumption for different number of points.]{
   \begin{tikzpicture}[scale = 0.53]
    \begin{axis}[name=plot1,height=6cm,width=8cm,
      legend style={at={(0.35,0.98)},{font=\footnotesize},anchor=north,legend columns=1}, enlargelimits=0.1,
      ybar=0pt, bar width=11,xtick align=inside, ymajorgrids, major grid style={dashed, line width=.7pt,draw=gray!50},
      xlabel={Number of points}, x label style={at={(0.5,-0.09)}}, 
      ylabel={Consumed energy (kJ)},y label style={at={(0.08,0.5)}},
      xtick={5,10,15,20,25,30},
    ]
    \addplot+[black!60,pattern=grid] 
    coordinates{
    (5,61.2749)
    (10,82.50653)
    (15,115.8994)
    (20,178.2028)
    (25,200.4815)
    (30,237.9840)
    };
    \addplot+[black!60, fill=black]
coordinates {
    (5,60.09644)
    (10,89.55519)
    (15,132.434)
    (20,162.653)
    (25,195.606)
    (30,231.669)
    };
    \vspace{-10pt}
    \legend{Field test,Prediction of simulator}
    \end{axis}
    \end{tikzpicture}
}
\caption{ The impact of the number of target points on the flight distance and flying time. }\label{fig:test-flight-m}
\end{figure}

By changing the the number of target points from 5 to 30, we obtain several different tasks. For each task, we generate a path planning and conduct simulation and field test accordingly. We record the flying distance 
and energy consumption in each task. 

As we can see from Fig.~\ref{fig:test-flight-m}
 , the flying distance
 and energy consumption in the simulation is very close to that in the field tests for multiple tasks.  In terms of the  energy consumption, the gap between  between the field test and simulator is  5.74\% on average.

\section{Related Work}
\label{sec:related}

This paper mainly focus on developing an easy-to-use unified framework to facilitate the design, deployment,  and testing of multi-UAV applications. In the introduction section, we have introduced most popular GCSs systems and differentiated our work from existing GCSs. In this section, we mainly review relevant multiple UAV cooperation testbeds and the existing work in path planning. 

Over the years, some research efforts have been made on designing testbeds~\cite{HKK-2004,BHL-2008,MMLK-2010,DCRDC-2015} for multiple UAVs. 
\nop{
How \textit{et al.}~\cite{HKK-2004} are among the earliest researchers to discuss testbed for UAV team cooperative control.
Baxter \textit{et al.} \cite{BHL-2008} design a multi-agent system to coordinate multiple UAVs in both simulated trials and test flights.
Michael \textit{et al.} \cite{MMLK-2010} address the development of a multiple micro-UAV testbed called GRASP to support research on coordinated and dynamic flight of MAVs.
In \cite{DCRDC-2015}, the researchers develop a number of tools for multiple fixed-wing UAVs supporting autopilot, collaborative autonomy, and human-swarm interface components.
}
Such testbeds for UAV are mainly focus on the control of fixed-wing UAV, which cannot be easily extended to that of multi-rotor UAV. In recent years, multi-rotor consumer UAVs, have become more and more popular. 
Multi-rotor UAV testbeds emerge quite recently. For example, the Group Autonomy for Mobile Systems (GAMS) project \cite{GAMS} aims at providing a distributed operating environment for accurate control of one or more UAVs. The OpenUAV project \cite{OpenUAV} provides a cloud-enabled testbed including standardized UAV hardware and an end-to-end simulation stack. 
Itkin \textit{et al.} \cite{IKP-2016} propose a cloud-based web application that provides real-time flight monitoring and management for quad-rotor UAVs to detect and prevent potential collisions. 
However, few works provide a generic platform that aims at simplifying the design of multi-UAV applications and enables the cooperation of multiple UAVs in a fleet. 
What is more, few works have exploited the connection characteristics among multiple  UAVs and thus constructed a mobile Ad-Hoc network to further enhance the robustness as well as quality of communications. 

In the past years, a number of research works~\cite{SB-2007, Maza-2007, barrientos-2011, KAYR-2015} have studies the path planning problem for coverage problem, however, they have not considered the connectivity requirement of multiple UAVs during the flight.
Only a few works have addressed the same connectivity requirement of UAV fleet during flight as considered in this paper. 
Yanmaz \etal \cite{Yanmaz-2012} and Schleich \textit{et al.} \cite{SJ-2013} develop and design distributed heuristic path planning algorithms to cover a given area and maintain the connectivity of UAVs, however, they fail to guarantee either the full coverage of the area or hard-constraint of connectivity.
Most recently, Bodin \etal~\cite{BCQS-IJCAI-2018} presented a demo that studies the path planning problem to visit a given set of points for cooperative connected UAVs.
However, this demo lacks details for the planning and only offline planning is presented, while read-world online synchronization and adjustment between UAVs during the flight is not considered.

In summary, to the best knowledge of the authors, there is a lack of viable solutions that can provide a generic framework/testbed to support cooperative UAV fleet control/monitoring with connectivity and  facilitate the deployment of multi-UAV applications.

\section{Conclusion}\label{sec:conclusion}

In this paper, we develop an open-source system named \systemname that enables the cooperation of multiple UAVs in a fleet. The proposed system provides generic interface that hides the hardware difference of UAVs to facilitate the cooperative UAV development, and also offers a series of cooperation services such as synchronization and connectivity management, energy simulation, and cooperative path planning to enable upper-layer user application designs. We further detail an application to demonstrate the services provided in our \systemname platform.  We evaluate  the system performance with simulations and field tests to validate that the proposed system is viable. 
A demo and source code are published in \textit{ https://github.com/whxru/CoUAV} for open access.

\ifCLASSOPTIONcaptionsoff
  \newpage
\fi

\bibliographystyle{IEEEtran}
\InputIfFileExists{couav.bbl}

\begin{IEEEbiography} [{\includegraphics[width=1in,height=1.25in,clip,keepaspectratio]{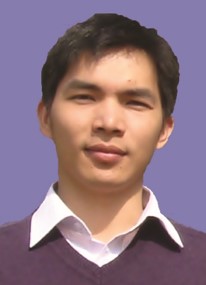}}]{Weiwei Wu}   is  an associate professor in Southeast University, P.R. China. He received his BSc degree in South China University of Technology and the PhD degree from City University of Hong Kong (CityU, Dept. of Computer Science) and University of Science and Technology of China (USTC) in 2011, and went to Nanyang Technological University (NTU, Mathematical Division, Singapore) for post-doctorial research in 2012.  His research interests include optimizations and  algorithm  analysis, wireless communications, crowdsourcing, cloud computing, reinforcement learning,   game theory and network economics.
\end{IEEEbiography} 
\begin{IEEEbiography}[{\includegraphics[width=1.1in,height=1.25in,clip,keepaspectratio]{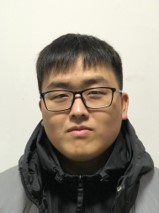}}]{Ziyao Huang} is currently a Master Student in Southeast University, P.R. China. He received his Bachelor's degree from Southeast University, P.R. China.  His research interests lie in the areas of  aerial swarm robotics, wireless communications, path planning, algorithm design.   
\end{IEEEbiography} 

\begin{IEEEbiography}[{\includegraphics[width=1in,height=1.25in,clip,keepaspectratio]{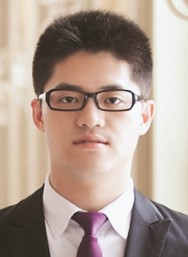}}]
{Feng Shan}  received his Ph.D. degree in Computer Science from Southeast University,   China in 2015. He is currently an Assistant Professor at School of Computer Science and Engineering, Southeast University. He was a Visiting Scholar at the School of Computing and Engineering, University of Missouri-Kansas City, Kansas City, MO, USA, from 2010 to 2012. His research interests are in the areas of energy harvesting, wireless power transfer, algorithm design and analysis.
\end{IEEEbiography}

\begin{IEEEbiography}[{\includegraphics[width=1in,height=1.25in,clip,keepaspectratio]{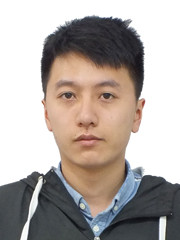}}]
{Yuxin Bian} is a Master Student in Southeast University, China. He received his Bachelor's degree from Nanjing University of Aeronautics and Astronautics.  His research interests lie in the areas of  wireless communications, schduling and reinforcement learning.   
\end{IEEEbiography}

\begin{IEEEbiography}[{\includegraphics[width=1in,height=1.25in,clip,keepaspectratio]{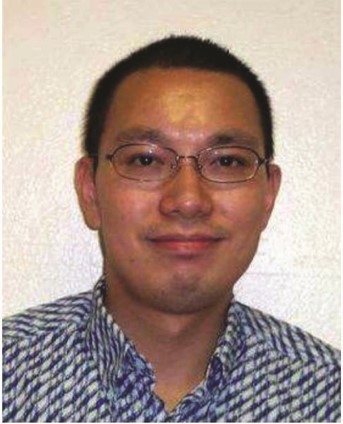}}]
{Kejie Lu}  
received the BSc
and MSc degrees in Telecommunications Engineering
from Beijing University of Posts and Telecommunications,
Beijing, China, in 1994 and 1997, respectively.
He received the PhD degree in Electrical
Engineering from the University of Texas at Dallas
in 2003. In 2004 and 2005, he was a Postdoctoral
Research Associate in the Department of Electrical
and Computer Engineering, University of Florida. In
July 2005, he joined the Department of Electrical
and Computer Engineering, University of Puerto
Rico at Mayag¨uez, where he is currently an Associate Professor. His research
interests include architecture and protocols design for computer and communication
networks, performance analysis, network security, and wireless
communications
\end{IEEEbiography}
\begin{IEEEbiography} [{\includegraphics[width=1in,height=1.25in,clip,keepaspectratio]{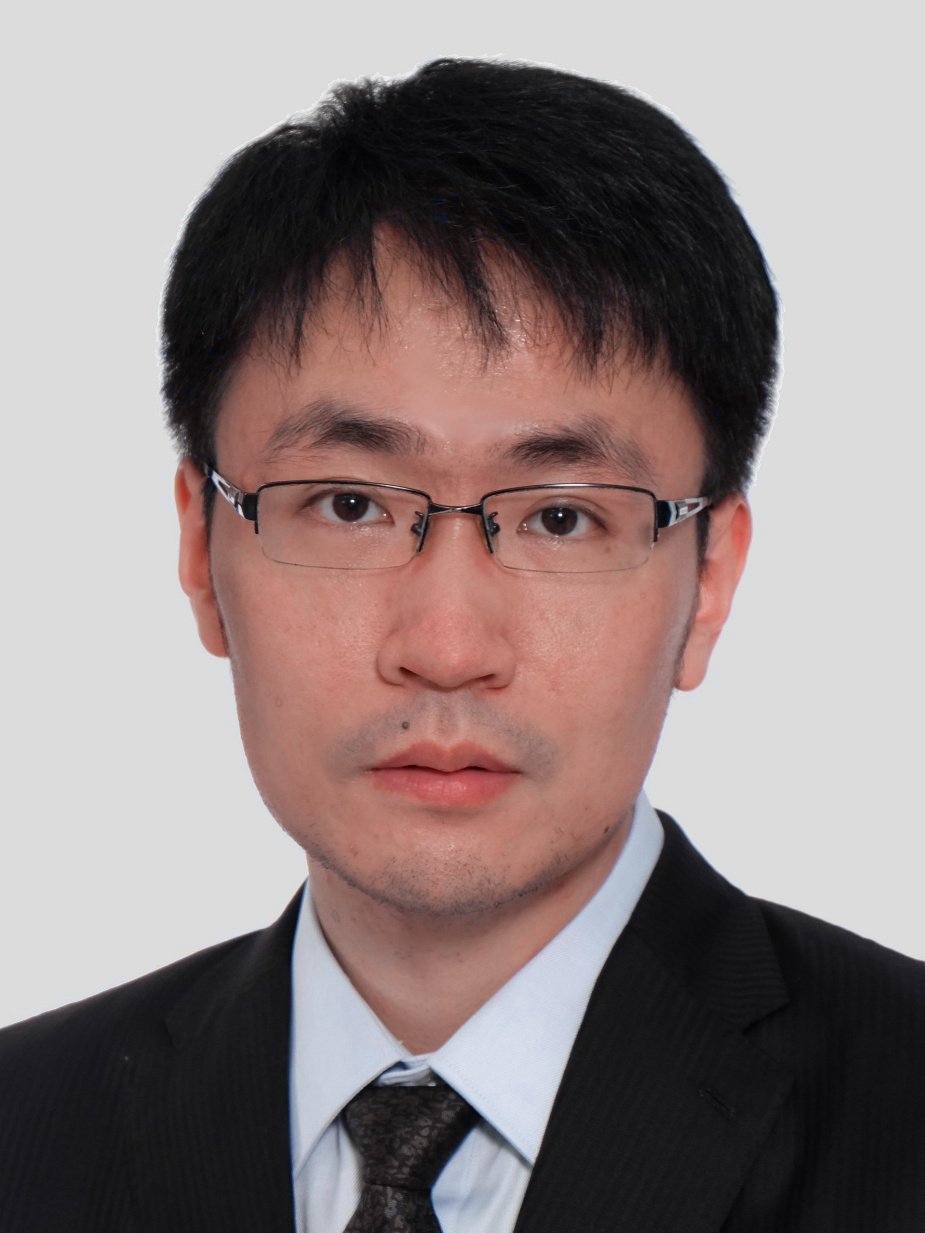}}]{Zhengjiang Li}  received the BE degree from Xian Jiaotong University, China, in 2007, and the MPhil and PhD degrees from the Hong Kong University of Science and Technology, Hong Kong, in 2009 and 2012, respectively. He is currently an assistant professor with the Department of Computer Science, City University of Hong Kong, Hong Kong. His research interests include wearable and mobile sensing, deep learning and mining, distributed, and edge computing. He is a member of the IEEE. 
\end{IEEEbiography} 
\begin{IEEEbiography} [{\includegraphics[width=1in,height=1.25in,clip,keepaspectratio]{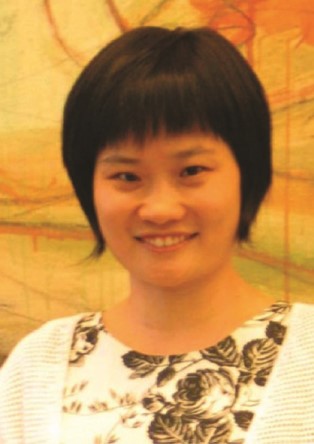}}]{Jianping Wang}  is an associate professor in the Department of Computer Science at City University of Hong Kong. She received the B.S. and the M.S. degrees in computer science from Nankai University, Tianjin, China in 1996 and 1999, respectively, and the Ph.D. degree in computer science from the University of Texas at Dallas in 2003. Jianping's research interests include dependable networking, optical networks, cloud computing, service oriented networking and data center networks. 
\end{IEEEbiography} \vfill

\end{document}